\documentclass[lettersize,journal]{IEEEtran}
\usepackage{amsmath,amsfonts}
\usepackage{graphicx}
\usepackage[backref]{hyperref}

\usepackage[dvipsnames]{xcolor}
\usepackage{algorithm}
\usepackage{listings}
\newcommand\algcomment[1]{\def\@algcomment{\footnotesize#1}}

\begin{document}

\title{Swin MAE: Masked Autoencoders for \\ Small Datasets}

\author{Zi'an Xu, Yin Dai, Fayu Liu, Weibing Chen, Yue Liu, Lifu Shi, Sheng Liu, Yuhang Zhou}

\maketitle

\begin{abstract}
The development of deep learning models in medical image analysis is majorly limited by the lack of large-sized and well-annotated datasets. Unsupervised learning does not require labels and is more suitable for solving medical image analysis problems. However, most of the current unsupervised learning methods need to be applied to large datasets. To make unsupervised learning applicable to small datasets, we proposed Swin MAE, which is a masked autoencoder with Swin Transformer as its backbone. Even on a dataset of only a few thousand medical images and without using any pre-trained models, Swin MAE is still able to learn useful semantic features purely from images. It can equal or even slightly outperform the supervised model obtained by Swin Transformer trained on ImageNet in terms of the transfer learning results of downstream tasks. The code is publicly available at \url{https://github.com/Zian-Xu/Swin-MAE}.
\end{abstract}


\section{Introduction}
\IEEEPARstart{I}{n} recent years, deep learning has seen rapid development. Particularly, after Transformer \cite{1} was proposed, there are more and more deep learning methods in the field of computer vision \cite{2, 3, 4, 5, 6, 7, 8, 9, 10}. However, further improvement of deep learning models in medical image analysis is majorly limited by the lack of large-sized and well-annotated datasets \cite{11}. Especially, the global prevalence of COVID-19 in recent years has put tremendous pressure on the healthcare industry, making it more challenging to collect large datasets and manually annotate them \cite{12, 13}.

As a deep learning method that does not require labels, unsupervised learning is more suitable for solving medical image analysis problems \cite{14, 15}. In recent years, unsupervised learning methods in the field of computer vision have also developed rapidly, and contrast learning methods such as BYOL \cite{16}, SimSiam \cite{17}, MoCo v3 \cite{18}, and DINO \cite{19} have achieved good unsupervised learning results in the field of natural images. However, all of these approaches strongly rely on data augmentation. Since medical images are mostly intact and single-channel, data augmentation methods commonly used in natural images (e.g., random cropping, color dithering) are not fully applicable to medical images \cite{20}, which makes it difficult to use these unsupervised learning methods in medical image analysis.

MAE is an unsupervised learning method proposed by Kaiming He in 2021 \cite{21}, which can learn useful semantic features purely from images without relying on data augmentation. In the year since MAE was proposed, there have been many excellent studies based on masked autoencoders in the field of natural images \cite{22, 23, 24, 25, 26}, but there are still fewer studies in the field of medical images \cite{27, 28}.

To make unsupervised learning applicable to small datasets, we proposed Swin MAE, which is a masked autoencoder with Swin Transformer \cite{29} as its backbone, as shown in Fig. 1. Compared to ViT \cite{30} used in MAE backbone, Swin Transformer introduces inductive bias similar to CNN, thus alleviating the problem of training Transformer dependent on large datasets and pre-trained models \cite{31}. Therefore, our Swin MAE can learn useful semantic features purely from images even on a dataset with only a few thousand medical images and without using any pre-trained models. It can equal or even slightly outperform the supervised model obtained by Swin Transformer trained on ImageNet in terms of transfer learning results of downstream tasks.

\begin{figure}[!t]
\centering
\includegraphics[width=3.5in]{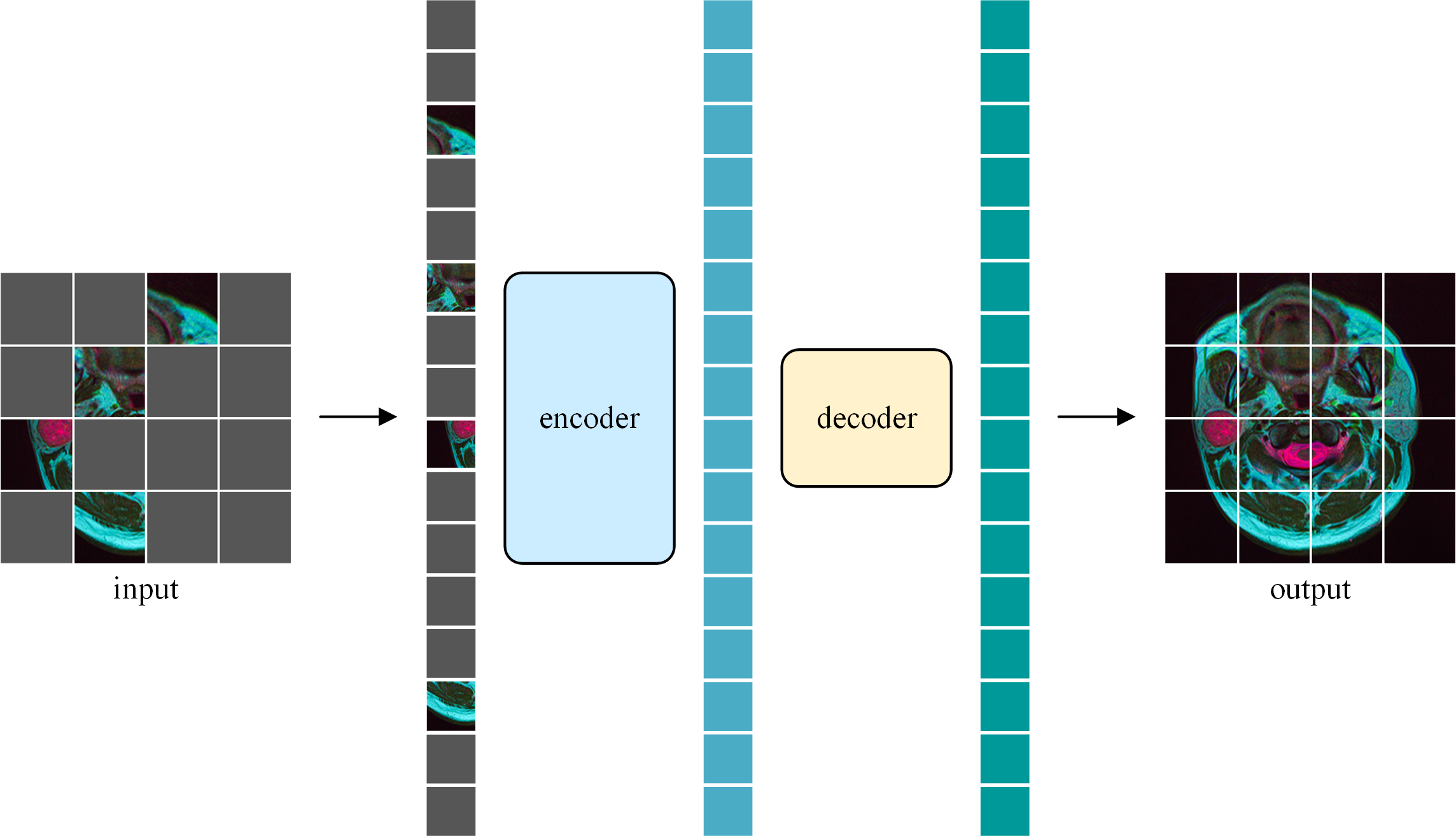}
\caption{Swin MAE schematic.}
\end{figure}

\begin{figure}[!t]
\centering
\includegraphics[width=3.5in]{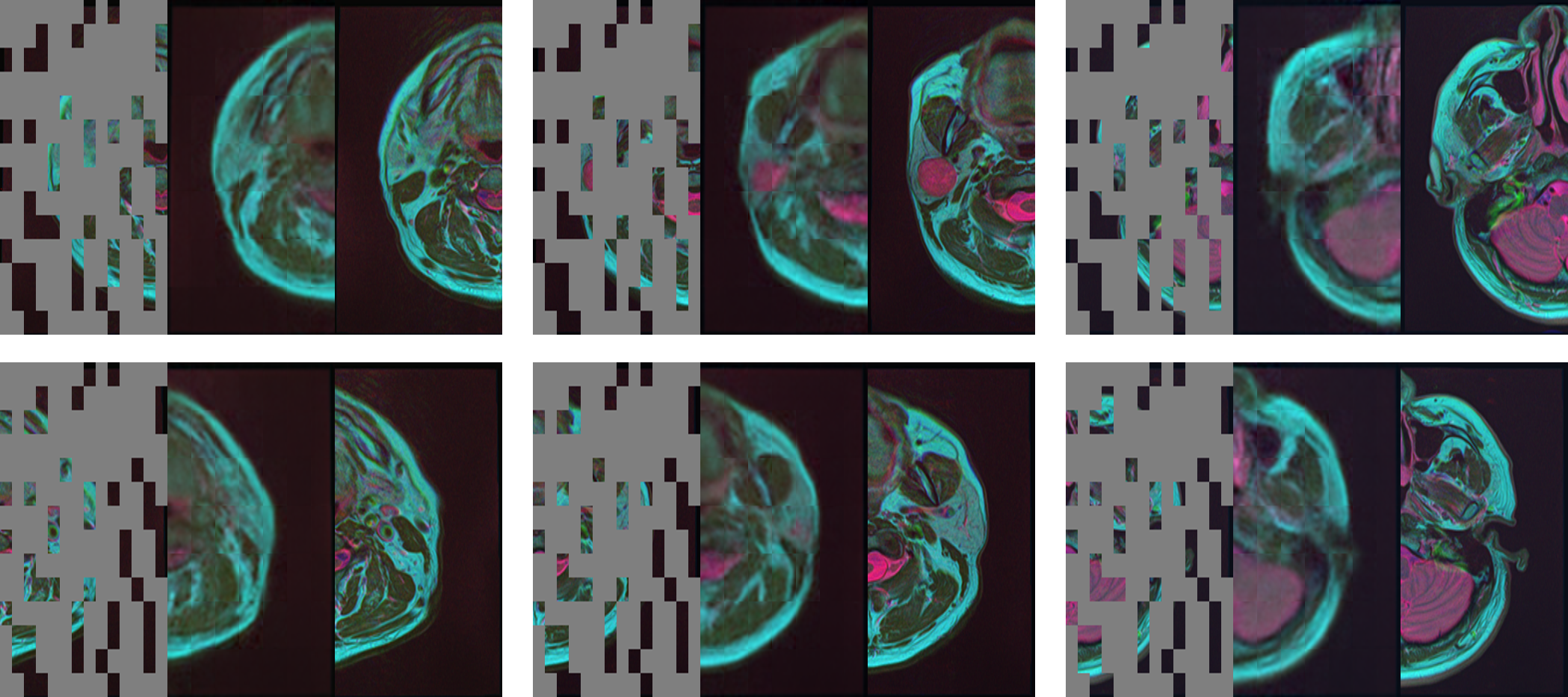}
\caption{Example results on validation images. For each triplet, we show the masked image (left), Swin MAE reconstruction (middle), and the ground-truth (right). The masking ratio used here is 0.75.}
\end{figure}

\begin{figure*}[!t]
\centering
\includegraphics[width=6in]{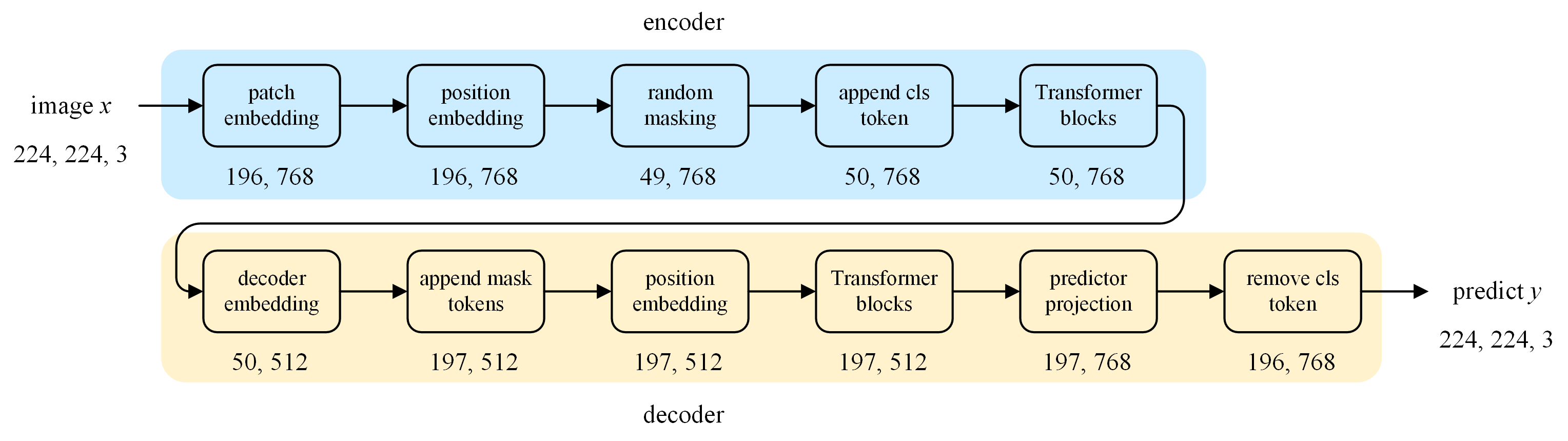}
\caption{MAE network structure.}
\end{figure*}

To visualize the role of the masked autoencoder and to observe the effect of unsupervised learning, the results of the trained Swin MAE applied to a part of the validation images are shown in Fig. 2. The color images in the figure are three-channel images displayed in RGB format, which is composed of three imaging sequences of MRI. In addition, during image preprocessing, the whole MRI image was divided into a left parotid image and a right parotid image from the middle. The image was resized to a square when it was input into the network. For presentation purposes, the image shown here is resized to the original aspect ratio, so the mask part of the image appears to be elongated.

It can be seen that Swin MAE complements the missing information in the masked image very well. Although in some details the reconstructions are slightly inconsistent with the real situation, the similarity of the reconstruction to the ground-truth is still striking. Specifically, many missing boundary contours are coherently filled in the reconstructed images, and the different colors of the missing regions in the images are well restored. This indicates that Swin MAE achieves good effect for all three imaging sequence images of MRI. In contrast, the texture details in the reconstructed image are lost to some extent, and the whole image looks blurrier, which is similar to the reconstruction results of MAE on natural images.

The rest of this paper is organized as follows. Section II describes the related work in this paper. Section III introduces the Swin MAE method. Section IV presents the relevant experiments and experimental results. Finally, our work is summarized in Section V.

\section{Related Work}
\subsection{MAE}
The Swin MAE proposed in this paper has many assumptions based on MAE \cite{21}. Similar to BERT \cite{32} in NLP, as an unsupervised learning method, MAE uses the masking learning approach. The input is obtained by masking the image randomly, and the output is the original image before being masked. Since the input and output of the network originate from the same object, the method is also a self-supervised learning method, and therefore its encoder is an autoencoder.

Since images do not contain individual words like sentences, to embed independent objects into tokens, as NLP does, the whole image needs to be artificially divided into independent patches. Traditional CNN networks require convolutional operations on the whole image and are therefore not suitable for doing so. The problem was solved with the introduction of ViT \cite{30}, and MAE follows ViT's approach to divide an image into regular non-overlapping patches. Therefore, it is logical that MAE also uses ViT as its backbone. Take the MAE obtained using the base version of ViT as an example, its network structure diagram is shown in Fig. 3.

There are some experimentally derived conclusions in the MAE paper, which are discussed briefly as follows:

\textbf{(i) Mask token.} Not removing the mask tokens during random masking not only increases the network computation but also makes the results worse. This may be because a large percentage of the encoder's input to the training is mask patches, which is not the case in real images. However, our Swin MAE does not remove the mask tokens and still has good performance in transfer learning of downstream tasks. The specific encoder design will be described in Section III.

\textbf{(ii) Reconstruction target.} Good results can be achieved by using original pixels instead of tokens as the reconstruction target. This means that the reconstruction result of the decoder does not have to be the same size as the output of the patch embedding layer, but only needs to satisfy the size that can be reshaped into the original image.

\textbf{(iii) Data augmentation.} Data augmentation brings little improvement to MAE, which indicates that MAE can learn useful semantic features purely from images without relying on data augmentation. Therefore, no data augmentation is used to train Swin MAE in the experiments of this paper.

\textbf{(iv) Mask sampling.} The effect of random sampling method is better than other regular sampling methods. Therefore, the masks in the Swin MAE experiments in this paper are also conducted in a random way. In order to solve the information redundancy caused by the small size of patches, the minimum masking unit in this paper is no longer a patch, but a window containing multiple neighboring patches. The window masking method will be described in detail in Section III.

\begin{figure*}[!t]
\centering
\includegraphics[width=5in]{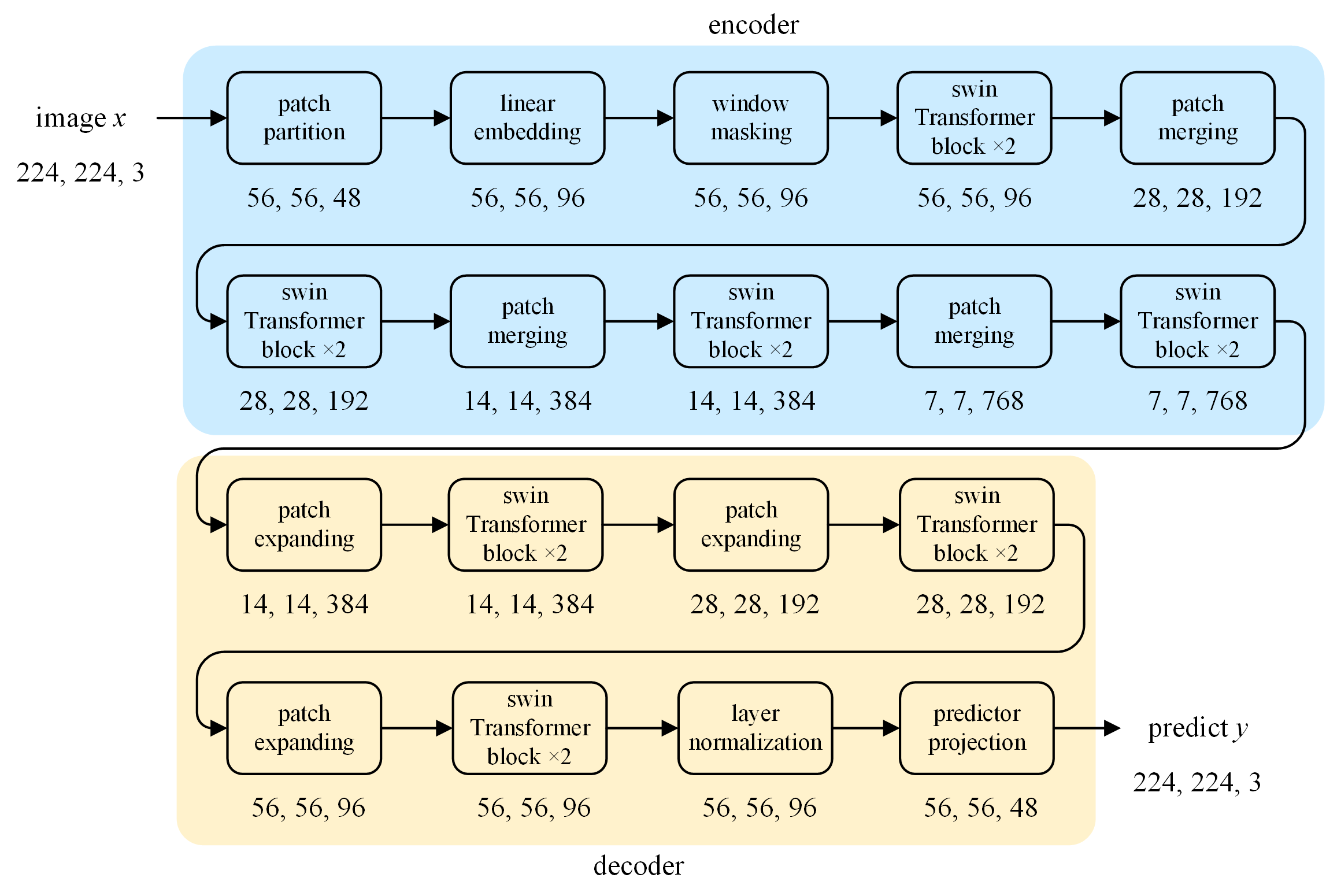}
\caption{Swin MAE network structure.}
\end{figure*}

\subsection{ViT and Swin Transformer}
ViT is the backbone of MAE, which has an efficient structure and good results. However, since ViT is based on Transformer, the network lacks the inductive bias of CNN (e.g., locality and translation equivariance). This makes training ViT very reliant on large datasets and pre-trained models \cite{31}. In the field of natural images, there are many datasets with millions of images such as ImageNet. However, in the field of medical images, most of the datasets are much smaller, making it more difficult to train ViT.

Swin Transformer \cite{29} uses a network structure similar to CNN. The patches are first divided using a 4$\times$4 size, followed by a Swin Transformer Block, and then a Patch Merging layer is used to reduce the number of tokens and expand the receptive field of each token. This introduces some of the inductive biases used in CNN, which reduces the dependence on the size of the dataset and gives better results in image analysis tasks. In addition, the network cannot compute the global attention due to the excessive patches divided in the first step, so Swin Transformer computes the window multi-head self-attention (W-MSA) instead. The exchange of information between patches belonging to different windows is achieved by computing the shifted window multi-head self-attention (SW-MSA). This allows Swin Transformer to have a computational complexity linearly related to the number of patches, which greatly improves the computational efficiency.

In summary, replacing the backbone of MAE with Swin Transformer will help to further improve the unsupervised learning effect on small datasets, which can improve the results of transfer learning for downstream tasks.

\section{Proposed Method}
\subsection{Overview}
Our Swin MAE is obtained by improving MAE \cite{21}, and the network structure is shown in Fig. 4. As a self-supervised learning method, similar to other autoencoder methods, Swin MAE can be divided into two parts: encoder and decoder. The encoder in the network is responsible for encoding the images into corresponding tokens and mapping them to a high-dimensional semantic space, while the decoder will refer to these latent representations and restore the original images. The loss function used for training calculates the mean square error of the original images and the reconstructed images. Only the mask patches are involved in the loss calculation, which makes the network training focus on predicting the mask patches. It should be noted that the decoder design shown in the figure is not fixed. The decoder of Swin MAE can be designed flexibly.

\subsection{Encoder design}
Following Swin Transformer \cite{29}, the whole image is divided into regular non-overlapping patches of 4$\times$4 size in the patch partition layer, and the length of tokens is later mapped to 96 by the linear embedding layer, which is the value used in Swin-T. The tokens are then continued into the subsequent four swin Transformer blocks in total, with no patch merging layer after the last block, consistent with the encoder of Swin-Unet \cite{33} used in the downstream task.

\begin{figure*}[!t]
\centering
\includegraphics[width=6in]{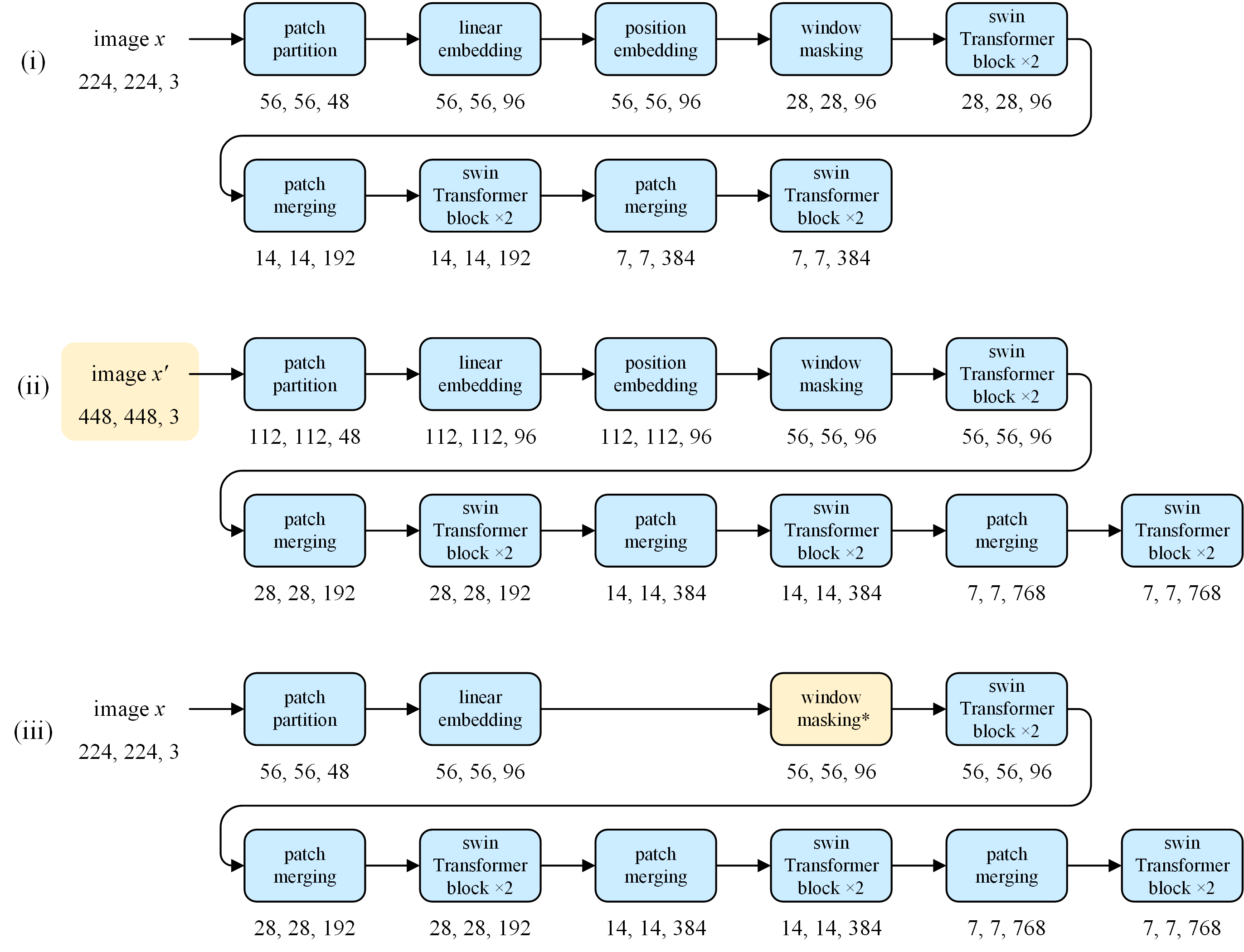}
\caption{Three different encoder designs.}
\end{figure*}

Unlike MAE, the window masking layer in Swin MAE does not remove the mask tokens, but directly and uniformly replaces them with a learnable vector, thus leaving the number of tokens unchanged. Removing these mask tokens will result in not enough tokens in the end to do the next patch merging. However, we experimented with three different encoder designs, as shown in Fig. 5, and described as follows: (i) Remove the final patch merging layer and swin transformer block directly. (ii) Double the length and width of the image so that the number of tokens obtained after embedding becomes four times the original number. After removing 75\% of all tokens that are masked, the number of tokens is the same as the original number. (iii) Do not remove the mask token. In addition, the relative position encoding is self-contained in the swin Transformer block, while using both relative and absolute position encoding does not theoretically lead to performance gains \cite{34}. If the mask tokens are not removed, there is no need to add the absolute position encoding before the window masking layer. Therefore, the position embedding layer is removed.

Each of the three encoder designs mentioned above has its own advantages and disadvantages. Encoder (i) causes the trained model to lack a layer of patch merging and a layer of swin Transformer block parameters when transferring to the downstream task, which affects the transfer learning result.

Encoder (ii) will not be missing the parameters of the patch merging layer and swin transformer block when transferring to the downstream task, but the parameters of the patch partition layer and the position embedding layer will not match the shape of the downstream network parameters. And while doubling the image length and width, the patch partition operation still uses a patch size of 4$\times$4, which results in each patch representing a quarter of the original image content. This further leads to a reduction in the information represented by the tokens obtained by embedding, which may affect the results of transferring to downstream tasks. Besides, the enlargement of image size will lead to an increase in the computation of network training and will also increase GPU memory usage.

Encoder (iii) does not remove the mask tokens. Although it is argued in the MAE paper that removing the mask tokens will instead lead to better training results. However, our experimental results show that design (iii) has the best results, so Swin MAE's encoder design uses this scheme, and the specific experimental data will be shown in Section IV.

\begin{figure*}[!t]
\centering
\includegraphics[width=6in]{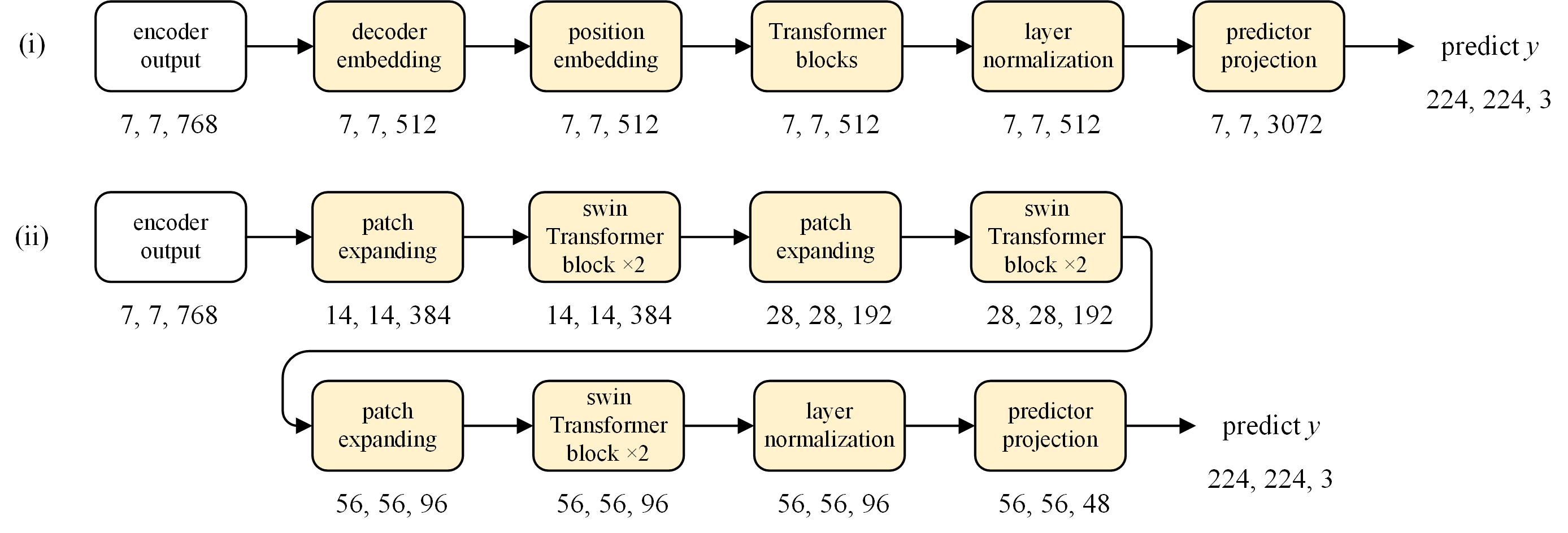}
\caption{Two different decoder designs.}
\end{figure*}

\subsection{Decoder design}
In general, the decoder weights are dropped after training, and only the encoder weights are used in transfer learning for downstream tasks, so the decoder can be designed flexibly. A lightweight decoder design will, on the one hand, reduce computation and memory usage, which can reduce training time and allow using a larger batch size, and on the other hand, make the network training more focused on the encoder, which leads to better transfer results on downstream tasks.

As mentioned in section II, since the reconstruction target of the decoder is the original pixels, it is fine as long as the final reconstruction size can be reshaped to the original image size. Suppose the original image height is $H$, the width is $W$, the number of channels is $C$, and the number of tokens reconstructed by the decoder is $L$. Then the dimension $D$ of the tokens is:

\begin{equation}
D = \frac{{H \cdot W \cdot C}}{L}
\end{equation}

Two different decoder designs are experimented with in this paper, as shown in Fig. 6, and described as follows: (i) The decoder uses ViT \cite{30} as the backbone and the structure is similar to the MAE decoder. (ii) The decoder uses Swin Transformer \cite{29} as the backbone and the structure is similar to the Swin-Unet decoder.

The decoder (i) is different from MAE in that the mask tokens are not removed in the encoder and the class token is not added, so there is no need to add the mask tokens nor remove the class token in the decoder. Meanwhile, since no mask token needs to be added, the number of tokens output by the decoder remains 7$\times$7, which is the same as the output of the encoder. Therefore, the dimension of the tokens which output by the decoder is determined by equation (1) as 3072.

The decoder (ii) differs from Swin-Unet in its predictor head part. Instead of using the final patch expanding layer to restore the original image dimensions, a predictor projection layer is directly used to map the dimensions of the tokens to 48 as MAE does. In addition, no skip connection layer is added between the encoder and decoder.

The experimental results show good results for both decoder (i) and (ii), which indicates that Swin MAE is not sensitive to the design of the decoder. Specific experimental data will be shown in Section IV.

\subsection{Window masking}
The masking method used in MAE is simple and efficient. The indexes to be masked are obtained by shuffling the index list and then taking the later part of the list according to the masking ratio. However, such a method uses a patch as the minimum unit for the mask operation. When the patch division size changes from 16$\times$16 used by ViT to 4$\times$4 used by Swin Transformer, this method causes some problems, as shown in Fig. 7 (b). Even with such a high masking ratio of 0.75, the small area of each patch still leads to the easy reconstruction of the mask patches using interpolation, which can easily lead Swin MAE to learn shortcut solutions. Therefore, the minimum unit of the random mask operation cannot be a single patch anymore, but rather a window containing multiple patches. We call this new masking method the window masking method. The mask results obtained using the new method are shown in Fig. 7 (c).

\begin{figure}[!t]
\centering
\includegraphics[width=3.5in]{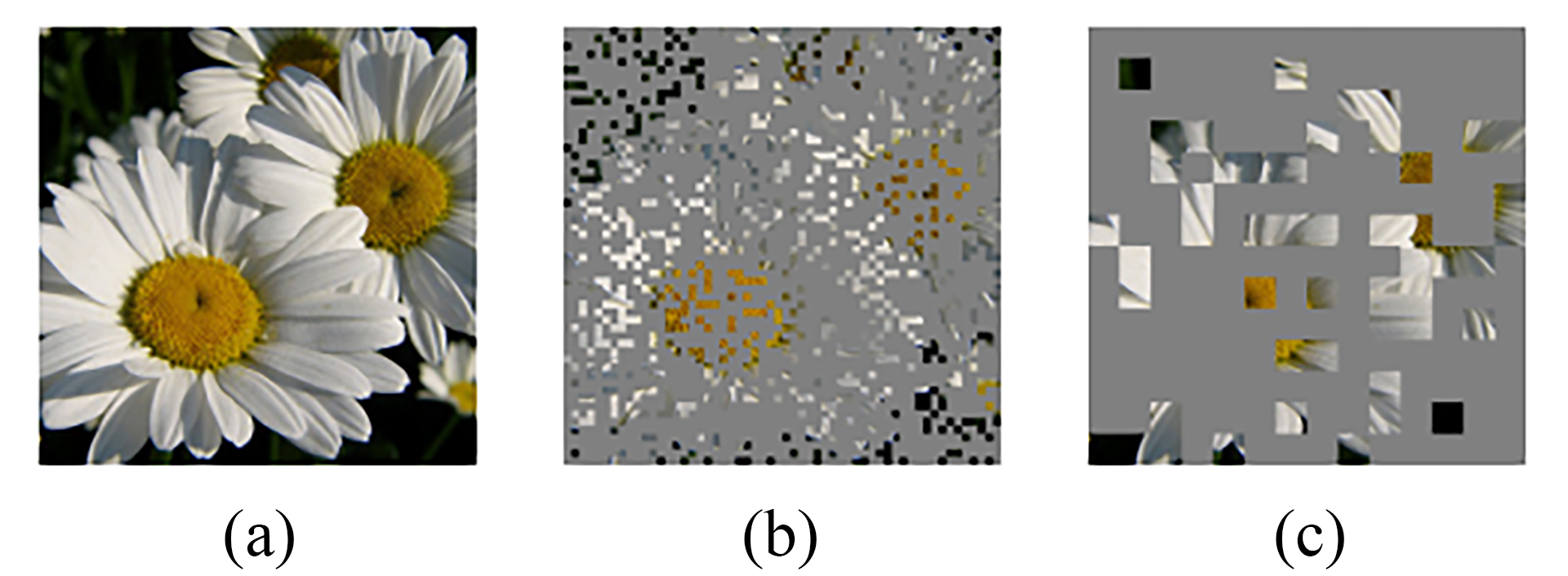}
\caption{The results of different masking methods, where (a) is the original image, (b) is the normal random masking method, and (c) is the window masking method. To facilitate the understanding of readers without specialized medical knowledge, a natural image is used here instead of the parotid MRI image for demonstration.}
\end{figure}

\begin{figure}[!t]
\centering
\includegraphics[width=3.5in]{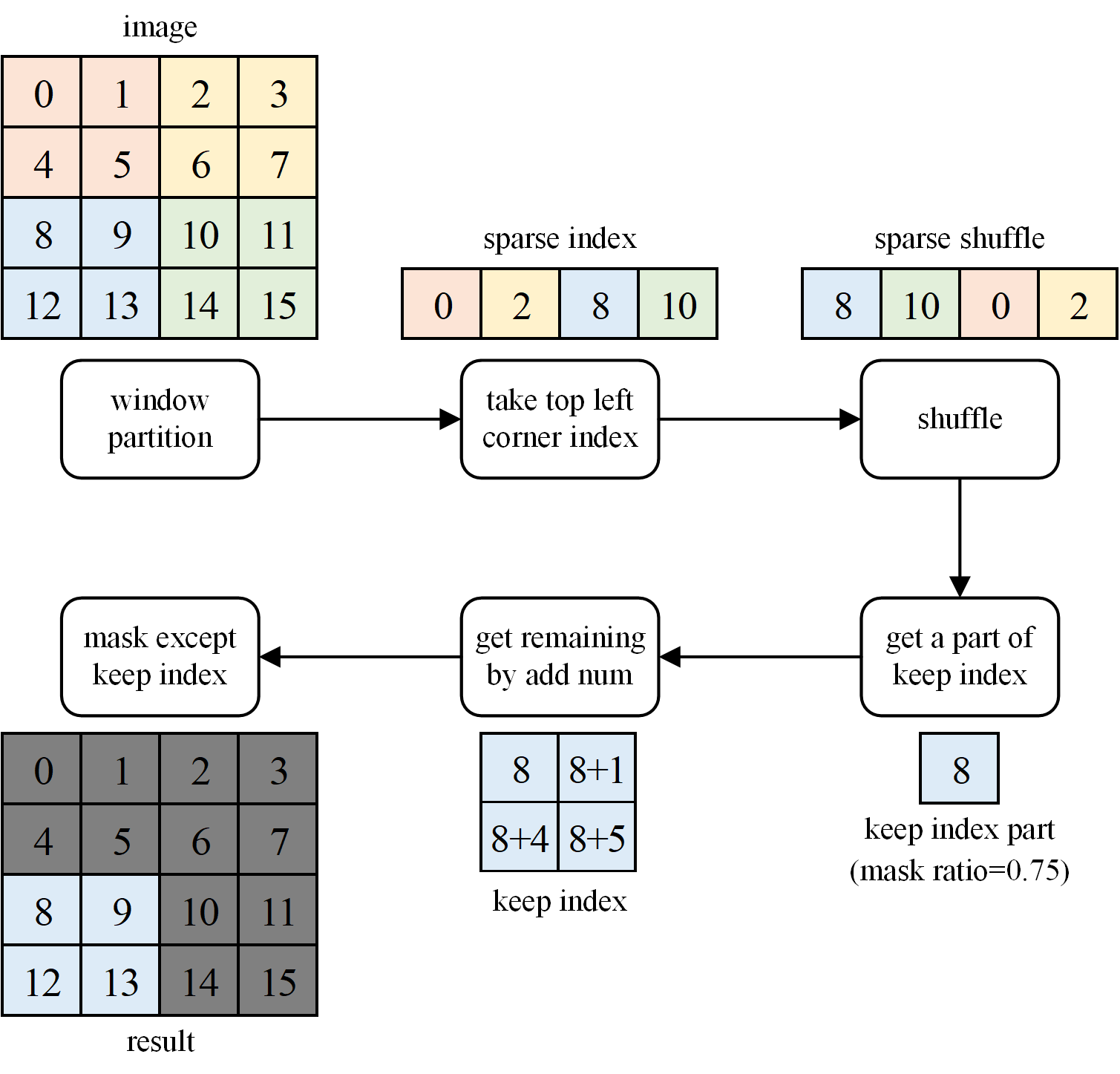}
\caption{Schematic diagram of window masking method.}
\end{figure}

\begin{algorithm*}[t]
\caption{Pseudocode of window masking method in a Python-like style.}
\label{alg:code}
\algcomment{\fontsize{7.2pt}{0em}\selectfont \texttt{bmm}: batch matrix multiplication; \texttt{mm}: matrix multiplication; \texttt{cat}: concatenation.
}
\definecolor{codeblue}{rgb}{0.25,0.5,0.5}
\lstset{
  backgroundcolor=\color{white},
  basicstyle=\fontsize{10pt}{10pt}\ttfamily\selectfont,
  columns=fullflexible,
  breaklines=true,
  captionpos=b,
  commentstyle=\fontsize{10pt}{10pt}\color{codeblue},
  keywordstyle=\fontsize{10pt}{10pt},
}
\begin{lstlisting}[language=python]
# d: d * d is the number of windows
# r: r * r is the number of tokens in a window
# mask_ratio: ratio of mask tokens to all tokens

# when mask_ratio > 0.5, the number of keep indexes is less than mask indexes, so finding keep indexes is more efficient

# obtain sparse indexes of keep tokens
noise = rand(d * d)  # noise in [0, 1]
sparse_shuffle = argsort(noise)  # get shuffled sparse indexes list
keep_ratio = 1 - mask_ratio  # ratio of keep tokens to all tokens
sparse_keep = sparse_shuffle[0 : d * d * keep_ratio]  # get sparse keep indexes list

# obtain indexes of a part of keep tokens by equation
index_keep_part = (sparse_keep // d) * d * r * r + (sparse_keep % d) * r

# obtain all indexes of keep tokens by add num
index_keep = []
for i in range(r):
	for j in range(r):
		index_keep.append(index_keep_part + d * r * i + j)
\end{lstlisting}
\end{algorithm*}

The practice of the window masking method is shown in Fig. 8. First, the two-dimensional array of token indexes is divided into regular non-overlapping windows, and the token indexes of the top-left corner of each window are taken and straightened to form a list. Since the indexes taken out are discontinuous, we call the composed list the sparse list. After shuffling the sparse list, the indexes that need to be kept are obtained according to the masking ratio in the same way as used in MAE. Since these indexes represent the tokens in the upper-left corner of the windows, the rest of the token indexes in the windows can be obtained by simply adding some numbers. Finally, we get the complete list of token indexes that need to be kept. Algorithm 1 provides the pseudocode of the window masking method for Swin MAE. The equation used in pseudocode can be derived in the following way.

Assuming that the lengths and widths of the original images are equal, the lengths and widths of the two-dimensional arrays composed of the token indexes are also equal. Let each window contains $r \times r$ tokens, and a total of $d \times d$ windows are divided, and the sparse index of one of the tokens in the sparse list is $x$. When the sparse list is turned into a two-dimensional array of $d \times d$, the two-dimensional coordinate of the index is $\left( {[x/d],x\bmod d} \right)$. The coordinate of this token in the two-dimensional array of all tokens is $\left( {[x/d] \cdot r,x\bmod d \cdot r} \right)$. Finally, after straightening, the index y of this token can be determined by:

\begin{equation}
y = [x/d] \cdot d \cdot {r^2} + (x\bmod d) \cdot r
\end{equation}

The experimental results show that the use of the window masking method can effectively improve the transfer learning performance of Swin MAE on the downstream task, and the specific experimental data are shown in Section IV.

In addition, the masking ratio used in the window masking layer is one of the hyperparameters that affect the transfer learning results of Swin MAE. A small masking ratio can make the task too simple and thus affect learning higher dimensional semantic features. Conversely, a large masking ratio can make the task too complex and thus lead to training underfitting. Therefore, ablation experiments using different masking ratios were designed in this paper. Based on the experimental results and the conclusions drawn in the MAE paper, we chose to set the masking ratio to 0.75, and the specific experimental data are also shown in Section IV.

\section{Experiments and Discussion}
Six different sets of ablation experiments were designed to test the effects of Swin MAE and to prove the conclusions mentioned in Section III. Default settings are briefly described below.

\textbf{Dataset.} The dataset used in the experiments in this section is the parotid dataset we collected. The dataset includes multicenter, multimodal MR images of 148 patients with parotid tumors. The parotid and tumor segmentation labels in the MR images were obtained by specialized clinicians. Each patient's MRI image contains three different imaging sequences of short time inversion recovery (STIR), T1-weighted sequence (T1), and T2-weighted sequence (T2). Following the approach of our previous study \cite{35}, the images of the same layers in the three imaging sequences were combined into three-channel images. For supervised learning of the downstream task, a total of 1897 labeled MR images were available. 80\% of these images are taken as the train set and the remaining 20\% as the test set. For unsupervised learning of the upstream task, there are 4688 unlabeled images in the dataset. Although the size of the dataset in the medical field is much smaller than the natural image dataset, Swin MAE still has a good unsupervised learning effect.

\textbf{Downstream tasks.} The Swin MAE proposed in this paper is an unsupervised learning method that requires a real downstream task to evaluate the actual transfer effect. We used parotid gland tumor segmentation as a downstream task. Specifically, the task is to perform semantic segmentation of regions of interest on MR images of the head and neck. The pixels in the image will be classified into three categories: (i) background, (ii) parotid gland, and (iii) tumor. Since the backbone of the encoder of Swin MAE is Swin Transformer, Swin-Unet \cite{33} with the same backbone is chosen as the segmentation network for the downstream tasks. The network structure of Swin-Unet is shown in Fig. 9. The batch size is set to 48, the maximum learning rate is set to $1 \times {10^{ - 4}}$, and 40 epochs are trained. As the number of training epochs increases, the learning rate decreases as a half-cycle cosine function, as shown in equation (3). Where $l{r_{\max }}$ is the maximum learning rate, $m$ is the total number of training epochs, and $i$ is the current number of training epochs. To alleviate the overfitting problems associated with small datasets, a variety of data augmentation methods are used on downstream tasks, specifically color dithering, gaussian filtering, horizontal flipping, and small angle rotation. To maintain the integrity of the MR images, data augmentation methods such as random scaling and cropping are avoided here.

\begin{equation}
lr = l{r_{\max }} \cdot (1 + \cos (\frac{i}{m} \cdot \pi ))/2
\end{equation}

\begin{figure}[!t]
\centering
\includegraphics[width=3.5in]{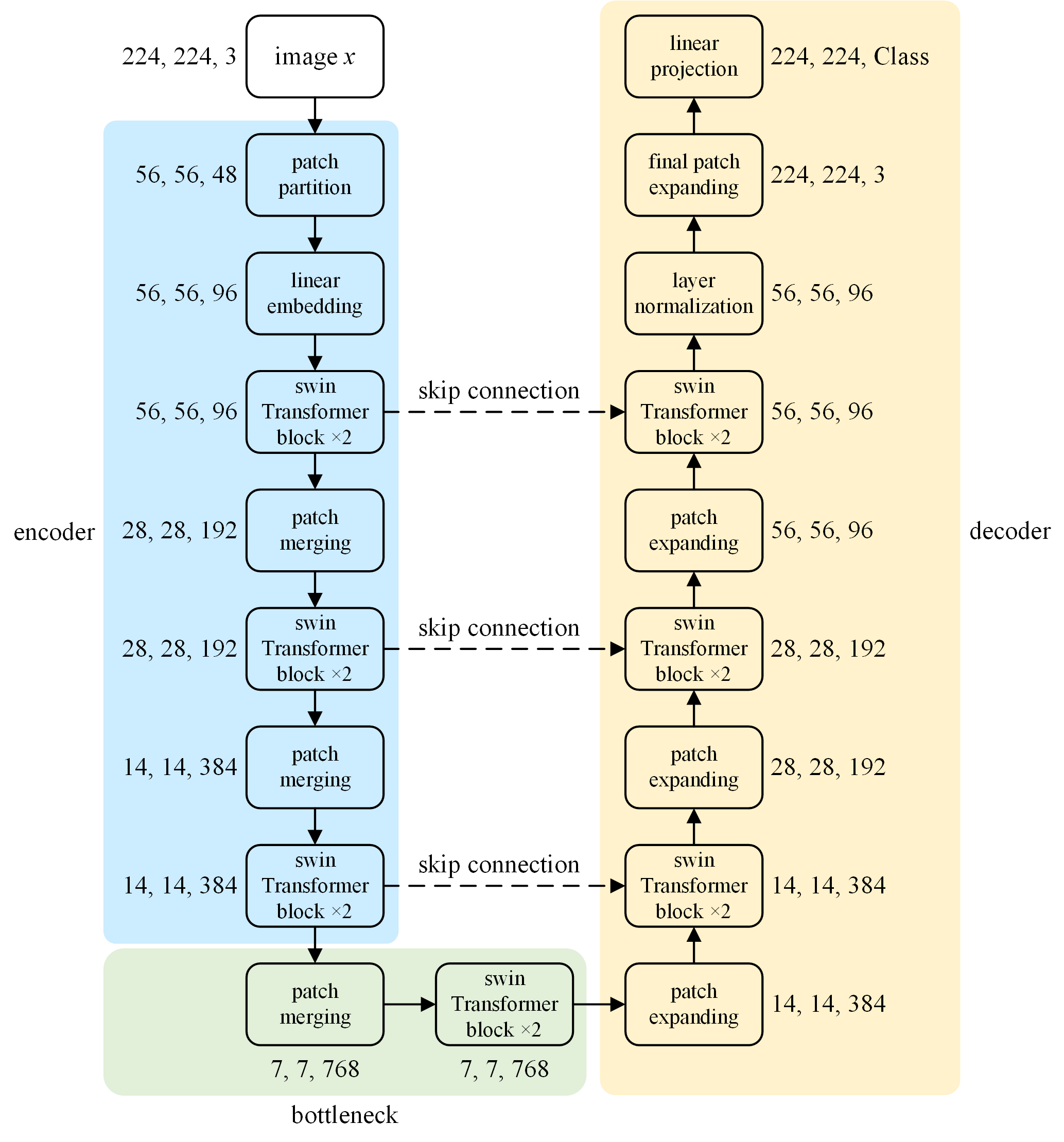}
\caption{Swin-Unet network structure.}
\end{figure}

\textbf{Transfer learning.} The Swin MAE model trained by unsupervised learning is used as a pre-trained model for transfer learning on downstream tasks. In general, the decoder weights of the Swin MAE model are dropped after training, leaving the encoder weights that can be used directly by the encoder and bottleneck in Swin-Unet. In the decoder of Swin-Unet, only the swin Transformer block uses the same pre-training weights as the corresponding layer in the encoder. The remaining patch expanding layer, skip connection layer and linear projection layer are trained using random initialization parameters. No parameters are frozen during training, and transfer learning is performed using fine-tuning of all layers.

\textbf{Evaluation metrics.} For the downstream segmentation task, we use four segmentation evaluation metrics to evaluate the results of the model on the test set. They are Dice-Similarity coefficient (DSC), Mean Pixel Accuracy (MPA), Mean Intersection over Union (MIoU), and Hausdorff Distance (HD), respectively. DSC, MPA, and MIoU respond to the area similarity of segmentation results, and larger values indicate better results. HD reflects the contour variability of the segmentation results, and smaller values indicate better results. The equations for the four metrics are as follows:

\begin{equation}
{\rm{ DSC }} = \frac{{2TP}}{{FP + 2TP + FN}}
\end{equation}

\begin{equation}
MPA = \frac{{TP + TN}}{{FN + TP + FP + TN}}
\end{equation}

\begin{equation}
MIoU = \frac{{TP}}{{FN + TP + FP}}
\end{equation}

\begin{equation}
HD(A,B) = \max (h(A,B),h(B,A))
\end{equation}

\subsection{Encoder design experiments}
In Section III, we designed three different encoder architectures, as shown in Fig. 5. We validated the effect of the three encoders using a set of ablation experiments, with the decoders all using ViT as the backbone. The decoder structures used in encoders (i) and (ii) are almost identical to the MAE decoder, except that class token is not used, as shown in Fig. 10. Encoder (iii) does not need to add the mask tokens in the decoder since the mask tokens are not removed. Therefore, the decoder used for encoder (iii) is shown as decoder (i) in Fig. 6.

\begin{figure}[!t]
\centering
\includegraphics[width=3.5in]{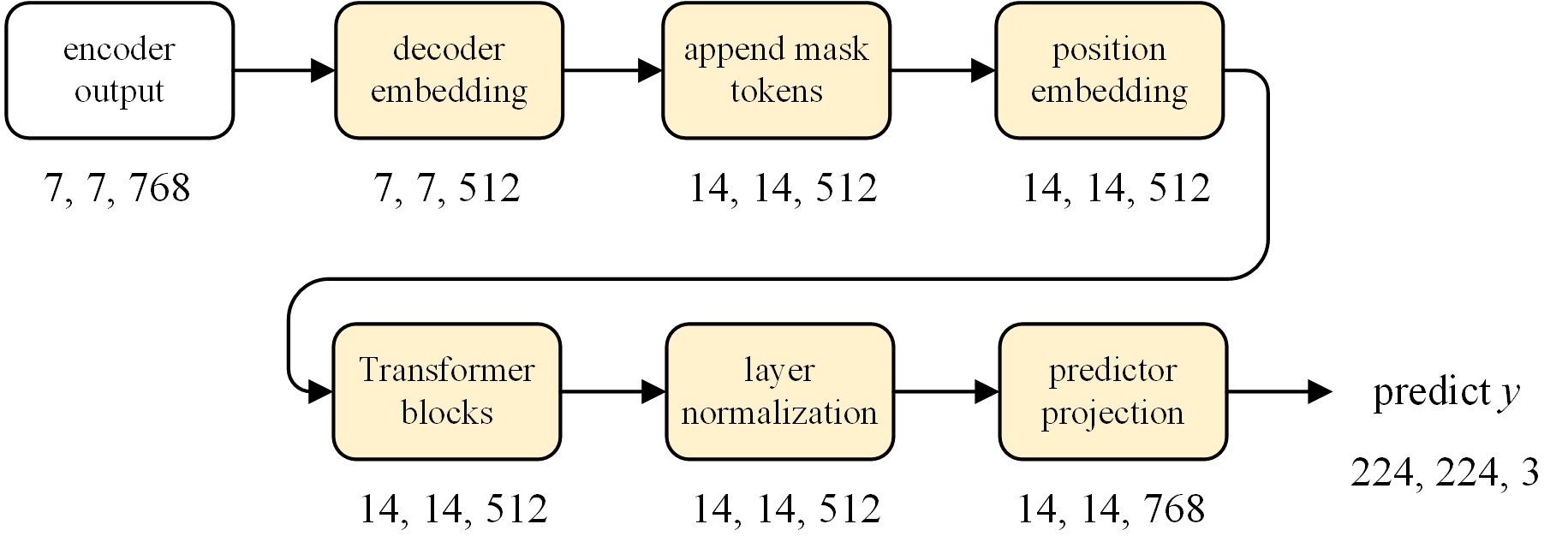}
\caption{Decoder used for encoder (i) and (ii).}
\end{figure}

The loss curves of Swin MAE composed of different encoders during the training of the upstream task are shown in Fig. 11. In terms of upstream tasks, encoder (iii) has the best training results. However, the real effect of unsupervised learning still needs to refer to the transfer learning results of downstream tasks.

\begin{figure}[!t]
\centering
\includegraphics[width=3.5in]{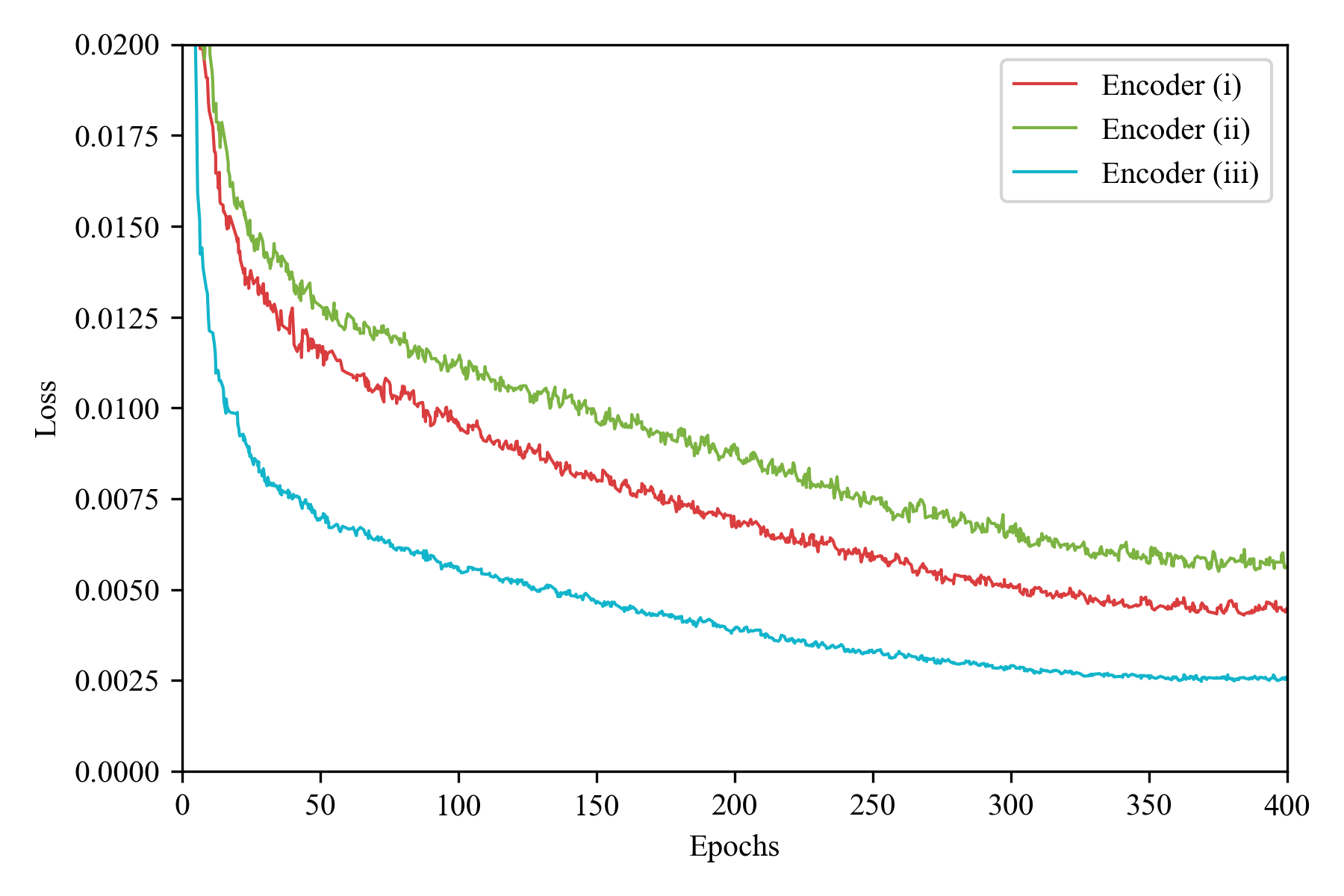}
\caption{Loss curves for encoder design experiment.}
\end{figure}

In designing the downstream task experiments, we consider that both encoders (i) and (ii) contain the position embedding layer, which is not included in the Swin-Unet used for the downstream task. This is a gap that may lead to poor transfer learning results of downstream tasks. Therefore, for encoders (i) and (ii), two networks with and without the position embedding layer are used for the downstream task. In addition, to verify whether the newly added position embedding layer affects the training effect of the network, experiments on training two networks with or without the position embedding layer by using random initialization are also included. For encoder (iii), since it does not contain the position embedding layer, only the network without the position embedding layer is used for the downstream task.

The MIoU curves of the test set during the downstream task training are shown in Fig. 12. The four evaluation metrics of the trained model on the test set are shown in Table I. Where None means that no pre-trained model is used, and training is performed with parameters initialized randomly. PE is the abbreviation for position embedding.

\begin{figure}[!t]
\centering
\includegraphics[width=3.5in]{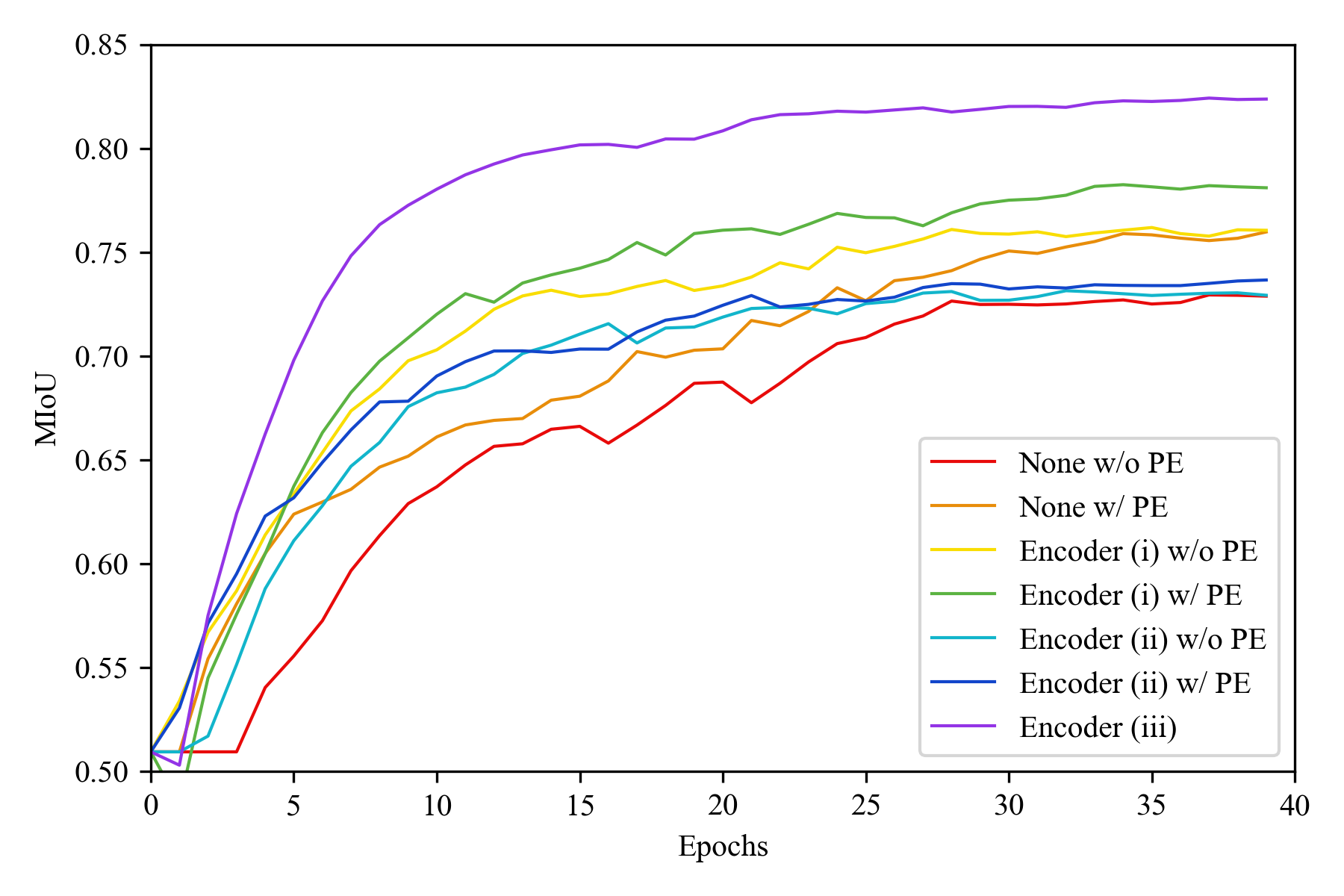}
\caption{MIoU curves in downstream tasks for encoder design experiment.}
\end{figure}

\begin{table}
\begin{center}
\caption{Transfer learning evaluation metrics for \\ encoder design experiment.}
\begin{tabular}{l c c c c}
\hline
Training method & DSC(\%) & MPA(\%) & MIoU(\%) & HD\\
\hline
None w/o PE & 79.45 & 98.88 & 73.47 & 3.60\\
None w/ PE & 81.91 & 99.02 & 76.17 & 3.43\\
Encoder (i) w/o PE & 82.11 & 98.94 & 76.18 & 3.51\\
Encoder (i) w/ PE & 84.03 & 99.05 & 78.44 & 3.39\\
Encoder (ii) w/o PE & 79.37 & 98.76 & 73.26 & 3.69\\
Encoder (ii) w/ PE & 80.01 & 98.83 & 74.13 & 3.63\\
\textbf{Encoder (iii)} & \textbf{88.03} & \textbf{99.23} & \textbf{82.96} & \textbf{3.13}\\
\hline
\end{tabular}
\end{center}
\end{table}

Experimental results show that adding absolute position encoding to Swin-Unet encoder instead leads to better results in this downstream task. Encoders (i) and (ii) have better results when absolute position encoding is added. Among them, encoder (i) achieves better results than None, which indicates that Swin MAE using encoder (i) also obtains some unsupervised learning effects. However, encoder (i) did not achieve the best results probably due to the missing part of the parameters of the encoder during transfer learning. Encoder (ii) does not improve the results compared to None due to the numerous problems analyzed in Section III. Encoder (iii) without removing the mask tokens achieves the best results, with more than 10\% improvement over None. Therefore, encoder (iii) is selected as the encoder design solution for Swin MAE.

\subsection{Decoder design experiments}
Section III mentioned that the decoder can be designed flexibly, and we tried two different decoder designs as shown in Fig. 6. We use a set of ablation experiments to verify the effect of both decoders. The encoders used in this experiment are all encoder (iii), which works best in the previous subsection.

Compared with the MAE decoder, decoder (ii) used in this experiment not only replaces the backbone with Swin Transformer but also removes the decoder embedding layer. The role of the decoder embedding layer in MAE is to change the output width 768 of the encoder into the input width 512 of the decoder, and its implementation is a fully connected layer. However, the output width of the Swin MAE encoder is as same as the input width of the decoder (ii), so there is no need to adjust the width. To verify whether the removal of this layer affects the effect of Swin MAE, we add a set of decoder (ii) with a decoder embedding layer to the ablation experiment. The loss curves of the training process for different decoders are shown in Fig. 13, where DE is the abbreviation of decoder embedding. It can be found that for different decoders, the training results of the upstream tasks are not significantly different.

As with many autoencoder methods, Swin MAE will generally only retain encoder weights when transferring to downstream tasks. However, decoder (ii) is almost identical to the decoder of Swin-Unet which is used for downstream tasks. If the decoder weights are also used as pre-training weights, will it have further improvement on the transfer learning results? To explore this question, we include an additional experiment using decoder (ii) and retaining decoder weights during transfer learning. The four evaluation metrics are shown in Table II, where DE is the abbreviation of decoder embedding, and DW is the abbreviation of decoder weights.

\begin{figure}[!t]
\centering
\includegraphics[width=3.5in]{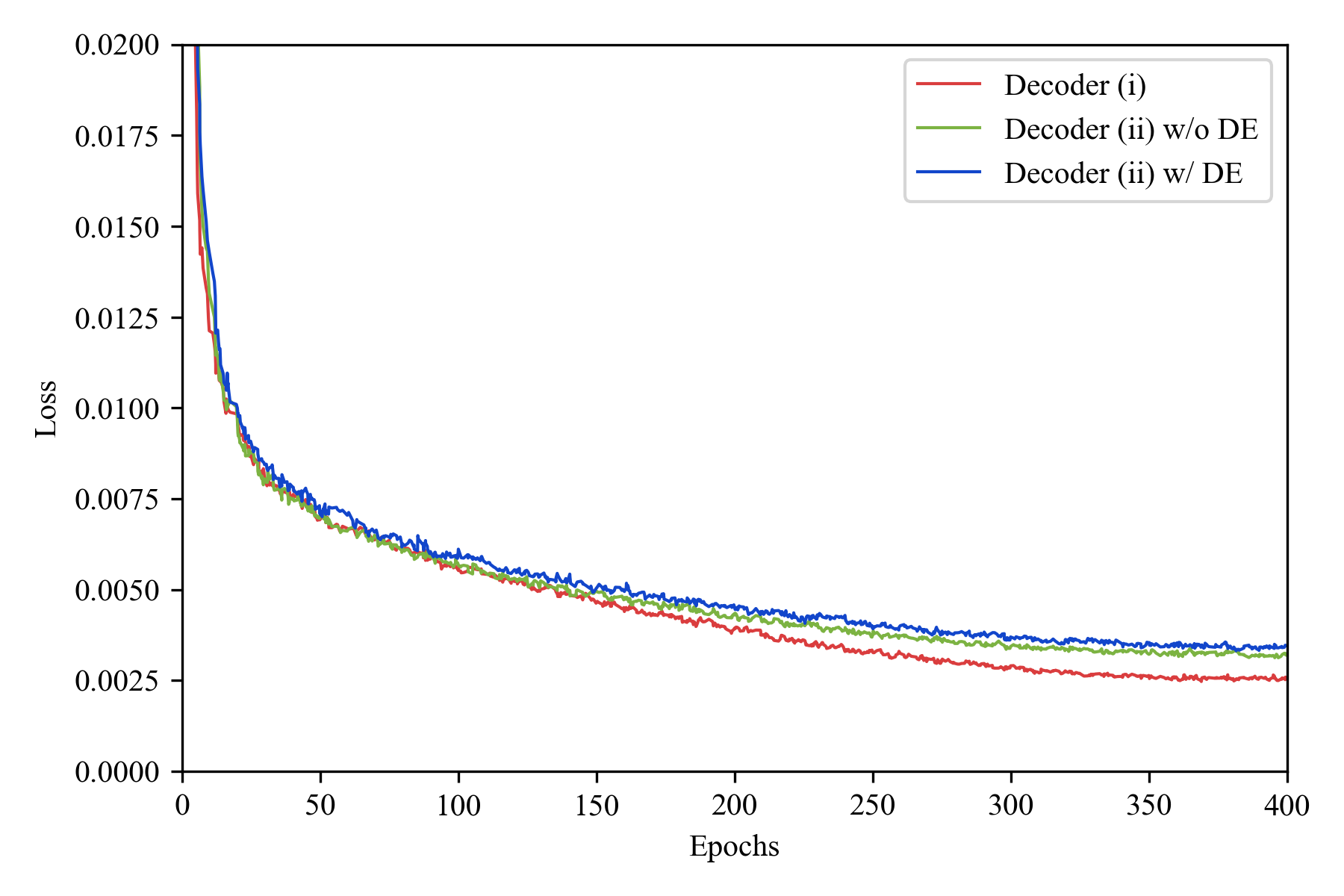}
\caption{Loss curves for decoder design experiment.}
\end{figure}

\begin{table}
\begin{center}
\caption{Transfer learning evaluation metrics for \\ decoder design experiment.}
\begin{tabular}{l c c c c}
\hline
Training method & DSC(\%) & MPA(\%) & MIoU(\%) & HD\\
\hline
Decoder (i) & 88.03 & 99.23 & 82.96 & 3.13\\
Decoder (ii) w/o DE & 87.60 & 99.22 & 82.44 & 3.15\\
Decoder (ii) w/ DE & 86.94 & 99.20 & 81.76 & 3.20\\
Decoder (ii) w/ DW & 87.37 & 99.21 & 82.17 & 3.16\\
\hline
\end{tabular}
\end{center}
\end{table}

The experimental results show that there is no significant difference between the results of decoder (i) and (ii), which indicates that Swin MAE composed by each decoder has good unsupervised learning effects. This indicates that the decoder design of Swin MAE can be flexible. Comparing the experimental results of adding and not adding decoder embedding layer, we can find that adding decoder embedding layer will make the training results worse instead. The reason may be that the added fully connected layer makes the decoder more complex, which increases the difficulty of model optimization. Therefore, if the output width of the encoder and the input width of the decoder is the same, the unnecessary decoder embedding layer can be removed. The results of the experiment in the last row of the table show that retaining and using decoder weights in transfer learning does not lead to better training results. We consider that this is due to the different task goals of the Swin MAE decoder used in the upstream task and the Swin-Unet decoder used in the downstream task. The former is to construct the original pixels of the masked image, while the latter is to predict the class to which the pixels belong. This is likely to be the gap that leads to poor decoder weight transfer learning results.

\subsection{Masking method experiments}
We propose the window masking method in Section III. To verify the effectiveness of the method, we designed a set of ablation experiments. Swin MAE in the experiments all use encoder (iii) and decoder (ii), the only difference is the different masking methods used. The loss curves of Swin MAE using different masking methods are shown in Fig. 14.

\begin{figure}[!t]
\centering
\includegraphics[width=3.5in]{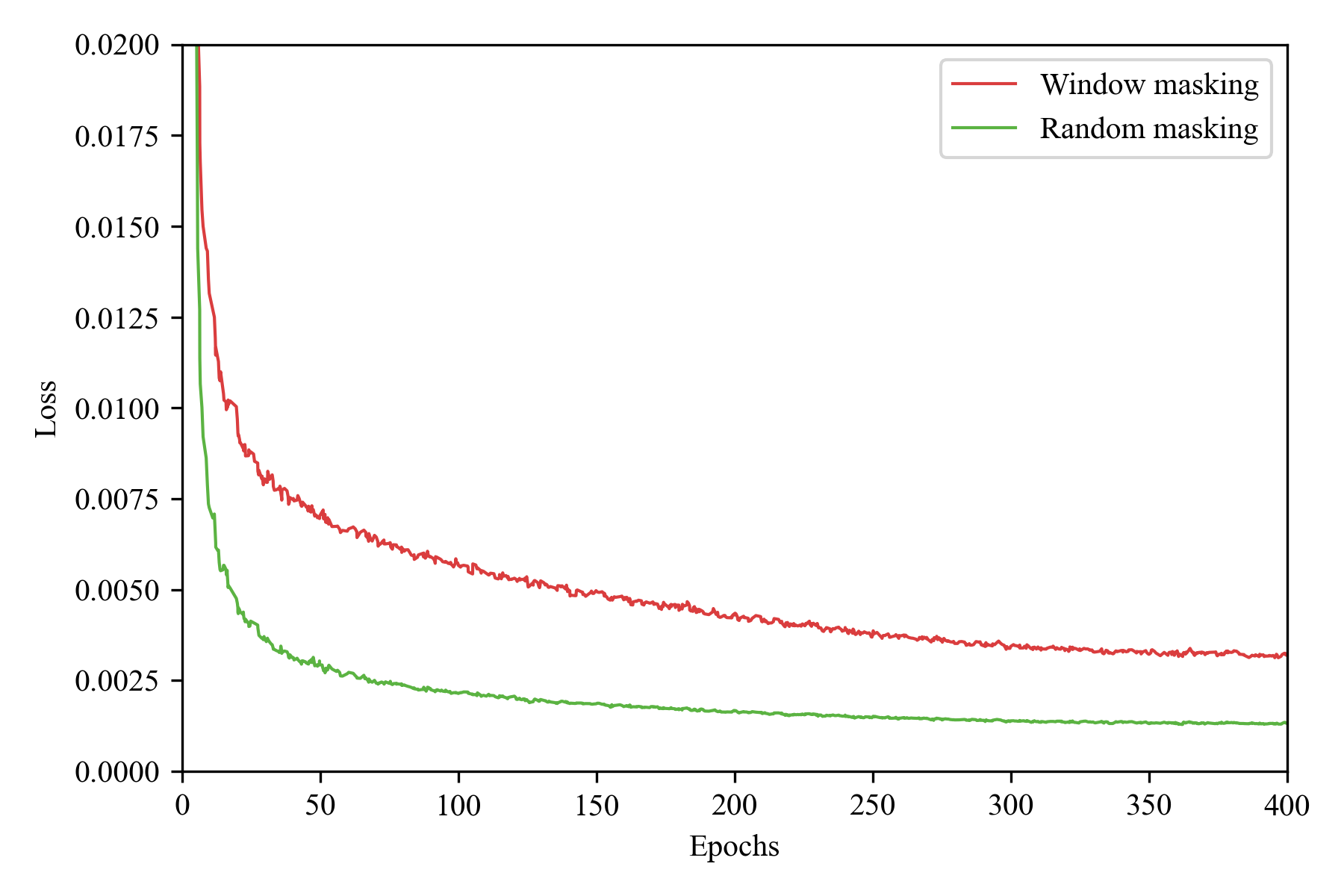}
\caption{Loss curves for masking method experiment.}
\end{figure}

Looking at the loss curve of the upstream task alone may suggest that the random masking method achieves better unsupervised learning results. But in fact, the lower loss of the random masking method is due to the low difficulty of its upstream task, as shown in Fig. 7(b), where the masking patches can be easily inferred from the neighboring patches. This is the reason why when evaluating the effectiveness of unsupervised learning, we need to take the transfer learning results of the downstream task as the criterion. The four evaluation metrics on the test set are shown in Table III.

\begin{table}
\begin{center}
\caption{Transfer learning evaluation metrics for \\ masking method experiment.}
\begin{tabular}{l c c c c}
\hline
Masking method & DSC(\%) & MPA(\%) & MIoU(\%) & HD\\
\hline
Random masking & 86.33 & 99.20 & 81.16 & 3.19\\
\textbf{Window masking} & \textbf{87.60} & \textbf{99.22} & \textbf{82.44} & \textbf{3.15}\\
\hline
\end{tabular}
\end{center}
\end{table}

The results show that although the random masking method has a lower loss for the upstream task, the transfer learning results of the window masking method are significantly better than the random masking method in the downstream task. Therefore, when the size of the patches divided by the patch partition layer is small, using the window masking method to mask multiple patches together can increase the difficulty of the upstream task and thus obtain better unsupervised learning results.

\subsection{Masking ratio experiments}
The masking ratio is the ratio of mask tokens to all tokens, and this hyperparameter affects the difficulty of upstream tasks. Upstream tasks that are too difficult or too easy are not conducive to Swin MAE learning useful latent representations, thus affecting the results of transfer learning for downstream tasks. Since medical images differ significantly from natural images, the applicable masking ratio may also be different. Therefore, we used ablation experiments to determine the most suitable masking ratio. Swin MAE uses encoder (iii) and decoder (ii) in the experiment, and the settings are the same except that the masking ratio is different. Fig. 15 shows the effect of the masking ratio on the transfer learning results of downstream tasks. Similar to the conclusion reached in the MAE paper, the optimal masking ratio of our Swin MAE is also around 0.7.

\begin{figure}[!t]
\centering
\includegraphics[width=3.5in]{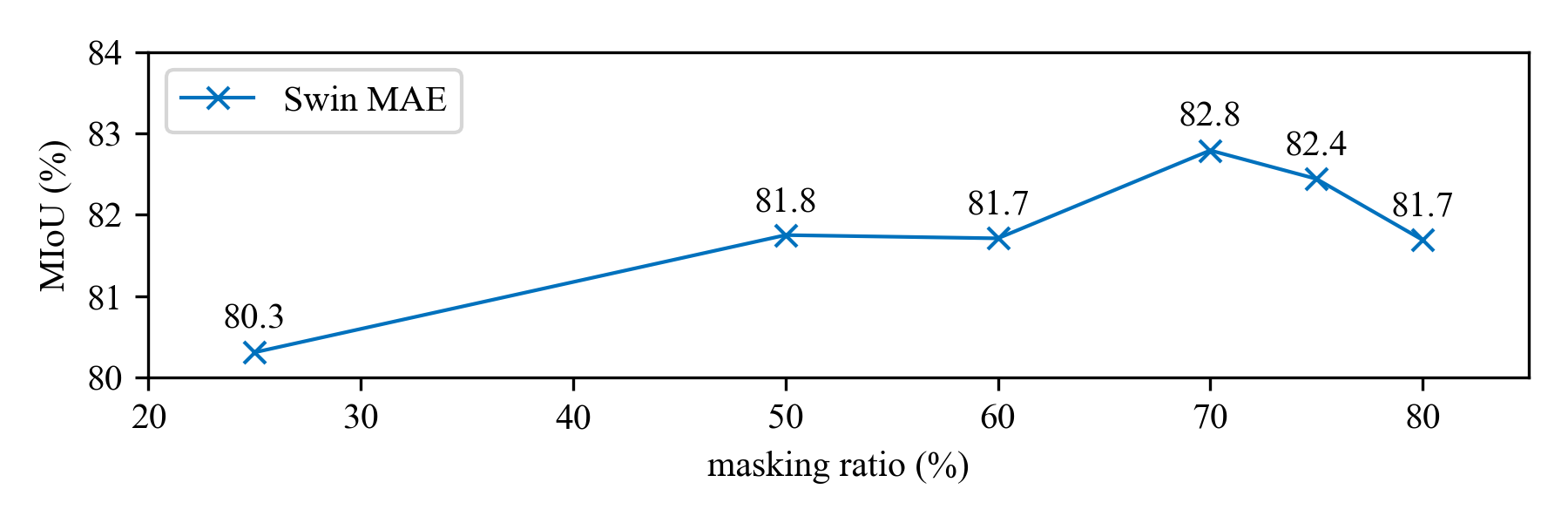}
\caption{MIoU line graph for downstream tasks using different masking ratio.}
\end{figure}

\subsection{Comparison with supervised learning}
In previous experiments, we compared the results of downstream tasks with random initialization and used different Swin MAE as pre-trained models. However, another common practice for transfer learning of downstream tasks is to use supervised learning models trained on the ImageNet dataset as pre-trained models. In this experiment, we will compare the transfer learning results of Swin MAE and supervised learning models. Swin MAE still uses encoder (iii) and the decoder uses both (i) and (ii). In addition, according to our previous research \cite{36}, using supervised learning models as initialization weights to train unsupervised learning networks may further improve performance. Therefore, we will also try to use the supervised models as pre-training weights for Swin MAE and observe whether its performance in transfer learning of downstream tasks will be further improved. The MIoU curves of the test set during the downstream task training are shown in Fig. 16. The four evaluation metrics of the model on the test set are shown in Table IV. Among them, None means no pre-trained model is used, Supervised means the supervised learning model trained on ImageNet is used, and S means the supervised learning model is used as the pre-trained model for Swin MAE.

\begin{figure}[!t]
\centering
\includegraphics[width=3.5in]{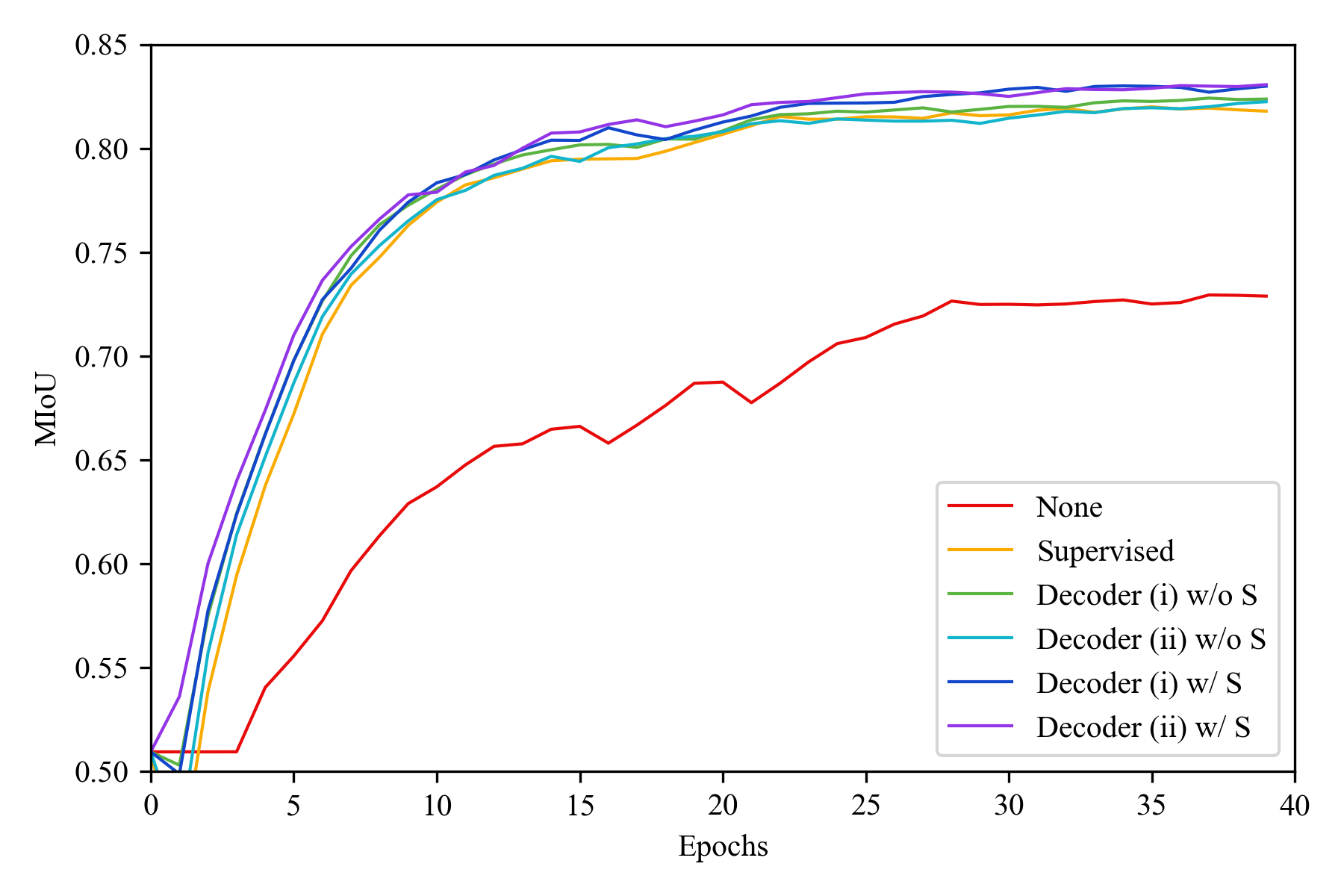}
\caption{MIoU curves of downstream tasks using different pre-trained models.}
\end{figure}

\begin{table}
\begin{center}
\caption{Transfer learning evaluation metrics of \\ downstream tasksusing different pre-trained models.}
\begin{tabular}{l c c c c}
\hline
Training method & DSC(\%) & MPA(\%) & MIoU(\%) & HD\\
\hline
None & 79.45 & 98.88 & 73.47 & 3.60\\
Supervised & 86.70 & 99.22 & 81.62 & 3.18\\
Decoder (i) w/o S & 88.03 & 99.23 & 82.96 & 3.13\\
Decoder (ii) w/o S & 87.60 & 99.22 & 82.44 & 3.15\\
Decoder (i) w/ S & 88.01 & 99.27 & 83.02 & 3.12\\
Decoder (ii) w/ S & 88.03 & 99.25 & 83.02 & 3.13\\
\hline
\end{tabular}
\end{center}
\end{table}

The experimental results show that even though the size of the medical dataset used is much smaller than that of ImageNet, the Swin MAE model trained with unsupervised learning on the medical dataset still exhibits good transfer effects and even slightly outperforms the model trained with supervised learning on ImageNet. However, the use of a supervised learning model as a pre-training model for Swin MAE just provides little improvement. This indicates that Swin MAE overcomes the common problem that Transformer networks are not easy to train, and obtains good results on small datasets without using any pre-trained weights. Therefore, Swin MAE is a good solution for the problem of unsupervised learning for small datasets.

\subsection{Comparison with MAE}
Transfer learning requires that the network structures used in the upstream and downstream tasks are approximately the same. Since MAE and Swin MAE have different backbones, the downstream network structures used for their transfer learning are also different. These differences make it difficult to fairly compare their transfer learning effects by designing ablation experiments in downstream tasks. To verify whether Swin MAE would have better accuracy than MAE on small datasets, we tried to design ablation experiments on the upstream task.

MAE divides the image into patches with a size of 16$\times$16 and uses one patch as the minimum masking unit. And Swin MAE divides the image into patches with a size of 4$\times$4. By using our window masking method to merge adjacent 4$\times$4 patches into one window and mask them together, the area of the minimum masking unit is the same. Then make both MAE and Swin MAE use a masking ratio of 0.75, so that the difficulty of the upstream tasks is the same. Same as MAE, Swin MAE uses the mean square error as the loss function and the evaluation metric for the upstream task. Since decoder weights are not involved in transfer learning of downstream tasks, to minimize the effect of different decoders, the Swin MAE in this experiment uses decoder (i), which is more similar to the decoder of MAE. With the same hyperparameters such as learning rate, the loss curves of MAE and Swin MAE for 400 epochs of training in the upstream task are shown in Fig. 17.

\begin{figure}[!t]
\centering
\includegraphics[width=3.5in]{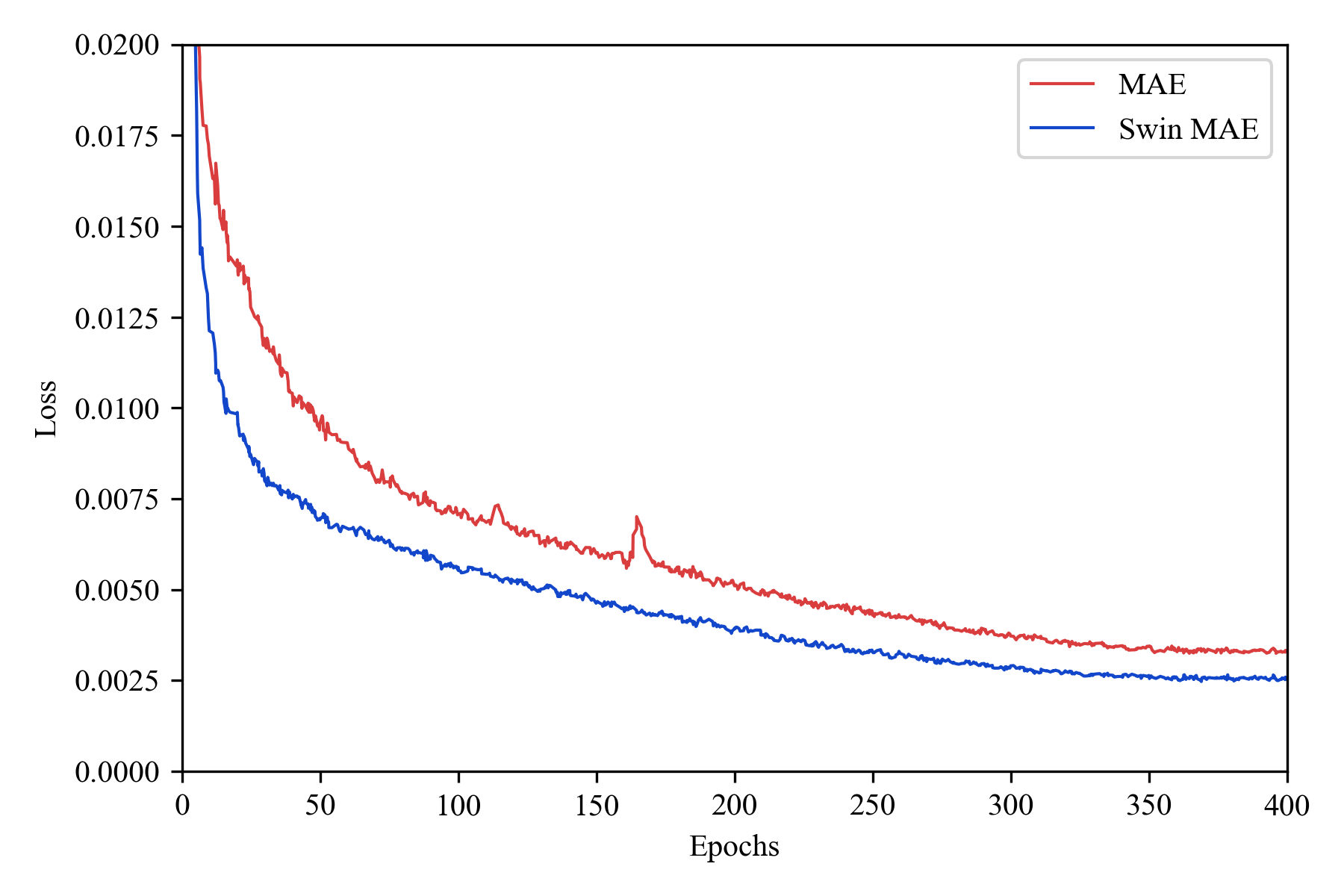}
\caption{The loss curves of MAE and Swin MAE.}
\end{figure}

Since the difficulty and evaluation metrics of the upstream task are the same, the loss curve of the upstream task can also roughly reflect the effect of unsupervised learning. The experimental results show that Swin MAE has (i) faster convergence, (ii) better training results, and (iii) smoother training curves compared to MAE on small datasets. This proves that Swin MAE is more advantageous for unsupervised learning tasks on small datasets.

\section{Conclusion}
Unsupervised learning for small datasets has always been a difficult problem. On the one hand, compared to supervised learning, unsupervised learning does not require labels to assist in training and only learns features from pure images, thus having a higher requirement for the size of the dataset. On the other hand, Transformer lacks the inductive bias contained in CNN, making it very dependent on pre-trained models during training. The Swin MAE proposed in this paper overcomes the problems mentioned above and can learn useful semantic features purely from images using only small datasets and without using any pre-trained models. This greatly improves the transfer learning results of downstream tasks, achieving equal or even slightly outperforming the supervised learning models trained on ImageNet.

Further, this means that deep learning does not only solve tasks that have large datasets and complete labels. In fact, more realistic tasks tend to have smaller datasets and costly manual labeling. An unsupervised learning method for small data sets can be very helpful for these tasks.

\textbf{Broader impacts.} Swin MAE predicts content based on the data used for training. The generated images represent the bias in the dataset and not necessarily the real situation. Unlike some image generation tasks of practical importance, such as image fusion and denoising tasks, the upstream tasks used by Swin MAE are only used to enable the network to learn useful semantic features, thus improving the transfer learning performance of the downstream tasks. There is no evidence that these inferred images can be used to solve clinical problems. These issues deserve further study and consideration based on this work.

\section*{Declaration of Competing Interest}
The authors declare that there is no conflict of interests.

\section*{Acknowledgments}
This research was funded in part by the Youth Program of National Natural Science Foundation of China grant number 61902058, in part by the Fundamental Research Funds for the Central Universities grant number N2019002, in part by the Natural Science Foundation of Liaoning Province grant number No. 2019-ZD-0751, in part by the Fundamental Research Funds for the Central Universities grant number No. JC2019025, in part by the Medical Imaging Intelligence Research grant number N2124006-3, in part by the National Natural Science Foundation of China grant number 61872075.

\bibliographystyle{IEEEtran}
\bibliography{paper}

\end{document}